\documentclass[conference]{IEEEtran}
\IEEEoverridecommandlockouts
\usepackage{cite}
\usepackage{amsmath,amssymb,amsfonts}
\usepackage{algorithmic}
\usepackage{graphicx}
\usepackage{textcomp}
\usepackage{xcolor}

\usepackage{caption}
\usepackage{subcaption}
\usepackage{pbox}
\usepackage{url}
\usepackage[boxed,ruled,linesnumbered,french]{algorithm2e}
\usepackage{bbold}
\usepackage{arydshln}

\usepackage{amsmath,amsfonts,bm}

\def\figref#1{Fig.~\ref{#1}}

\def\secref#1{section~\ref{#1}}
\def\Secref#1{Section~\ref{#1}}

\def\eqref#1{equation~\ref{#1}}

\def\1{\bm{1}}

\def\vx{{\bm{x}}}

\def\evx{{x}}

\DeclareMathAlphabet{\mathsfit}{\encodingdefault}{\sfdefault}{m}{sl}
\SetMathAlphabet{\mathsfit}{bold}{\encodingdefault}{\sfdefault}{bx}{n}

\begin{document}

\title{Learning Dynamic Author Representations with Temporal Language Models\\
\thanks{This work has been partially supported by the ANR (French National Research Agency) LOCUST project (ANR-15-CE23-0027)}
}

\author{
    \IEEEauthorblockN{Edouard Delasalles, Sylvain Lamprier}
    \IEEEauthorblockA{\textit{Sorbonne Universit\'e, LIP6, F-75005, Paris, France} \\
    edouard.delasalles@lip6.fr, sylvain.lamprier@lip6.fr}
    \and
    \IEEEauthorblockN{Ludovic Denoyer}
    \IEEEauthorblockA{\textit{Facebook AI Research} \\
    denoyer@fb.com}
}

\maketitle

\begin{abstract}
Language models are at the heart of numerous works, notably in the text mining and information retrieval communities. These statistical models aim at extracting word distributions, from simple unigram models to recurrent approaches with latent variables that capture subtle dependencies in texts. However, those models are learned from word sequences only, and authors' identities, as well as publication dates, are seldom considered. We propose a neural model, based on recurrent language modeling, which aims at capturing language diffusion tendencies in author communities through time. By conditioning language models with author and temporal vector states, we are able to leverage the latent dependencies between the text contexts. This allows us to beat several temporal and non-temporal language baselines on two real-world corpora, and to learn meaningful author representations that vary through time.
\end{abstract}


\section{Introduction}
Language modeling has been at the heart of a huge amount of works for decades. While the natural language processing field focuses on fine-grained text analysis, statistical models for information retrieval and text mining are essentially based on word (or N-gram) counts, considering more or less complex dependencies in texts. Early works in this area focused on the unigram multinomial model \cite{song99}, and recent works are shifting toward neural approaches, with distributed representations of words \cite{bengio2003, mikolov10}. Research on these deep language models is very active \cite{vaswani17,merity2017regularizing,bai18,melis18,merity18analysis}, with applications in various text-related tasks such as speech recognition \cite{chiu17}, image captioning \cite{vinyals17}, or text generation \cite{fedus18}.

The goal of the language modeling task is to determine word distributions, depending on their context. Classically, these contexts are limited to previous or surrounding words in text documents. However, textual documents often come with additional contextual information, namely their authors and publication dates. Leveraging this additional contextual information is thus a key challenge in order to build more efficient language models.
While the Authorship Analysis domain focuses on modeling author's writing style \cite{ding2017learning}, very few works focus on the combined consideration of the author and the publication date of textual documents. However, language changes through time, and authors style, as well as their writing subjects, change too.
It is in the domain of information diffusion, which studies content transmissions in information networks \cite{saito09}, that most of the work on dynamic extraction and prediction of relationships between authors through time has been proposed. However, almost all of the proposed approaches focus on the study of the information spread in a binary setting (infection or non-infection by a content emitted from one source in the network). Now, it appears obvious that dynamics in author communities (inter-author influences or patterns of reactions to some external stimuli) are not limited to binary events, but are also reflected in more diffuse behaviors, and notably on the way people communicate. Various works on topic modeling and their temporal evolution exist \cite{wang06, kaban02}, but they do not consider the multi-authors setting. Moreover, they are built on bag-of-words representation, and thus cannot directly leverage the representation learning power of deep language models.
 
We propose to study language evolution dynamics in author communities from a deep language model perspective. We establish a dynamic model of the language evolution in an author community based on representation learning. Our model is able to capture latent dynamics in the community via a combination of static and dynamic author representations. The dynamic representations are updated at each timestep with a residual transition model. These states condition a deep language model, enabling it to take into account temporal trends among authors. This state-based conditioning is similar in spirit to \cite{le14}, where a variation of the \texttt{Word2Vec} model \cite{mikolov2013} is conditioned on the paragraph from which the considered text is extracted. We conducted experiments on a scientific publications corpus and a news corpus for several temporal tasks: modeling (all timesteps are visible), imputation (random timesteps are hidden), and prediction (future timesteps are hidden). Our method consistently achieves state of the art performance on all tasks. Moreover, we performed quantitative and qualitative studies of the learned latent representations and show that our model is able to learn meaningful representations.

The remaining of this paper is organized as follows. In \secref{sec:rw} we present the related work. \Secref{sec:model} details our approach. Finally, \secref{sec:setup} describes our experimental protocol and \secref{sec:xp} details the results.

\section{Related Work}
\label{sec:rw}

Interest for language evolution in texts is not new. Going back some fifteen years, we find work on the evolution of topics in textual documents, notably  \cite{kaban02} whose model based on hidden Markov chains seeks to visualize temporal evolution in a textual stream. This approach falls in the general field of \textit{Topic Detection and Tracking}, where the idea is to identify and follow trending topics in streams. The approach, which extends the GTM temporal model of \cite{bishop97} for textual modeling, allows one to visualize the thematic changes via trajectories on a two-dimensional grid. However, this kind of work enables to track thematics and to segment texts but cannot be used for language modeling. The non-markovian approach proposed in \cite{wang06} is restricted to bag-of-words representations, but has a good ability to detect the topics' evolution over the observation period. Besides, various works studied temporal vocabulary evolution - according to semantic graph transformations in  \cite{kenter15} -, or thematic shifts in author communities - according to the dominant topics per time-step in \cite{hall08}.

Closer to applications targeted in this paper, dynamic topic models \cite{blei06} propose an \texttt{LDA}-like modeling (Latent Dirichlet Allocation \cite{blei02}), where the topic distributions and the distributions of words w.r.t. topics evolve over time. The evolution between successive multinomial distributions are driven by Gaussian motions of their natural parameters, in a Kalman filters fashion, and optimized via variational inference. However, these approaches require manually setting the number of topics, and language models are limited to simple word occurrence distributions. It is not trivial to include models with long-term dependencies, such as LSTM, in this context. Moreover, contrary to ours, these approaches are usually constrained to specific conjugate distributions for the inference of the latent variables of their evolution model. Note the extensions of \cite{blei06} to a multi-scale temporal version \cite{iwata12} or to a model with continuous-time dependencies  \cite{wang12}. Besides, \cite{gerrish10} introduces the concept of influence between documents, which could get closer to our objective but which is limited to analysis tasks. Lastly, \cite{wang11} proposes a temporal approach which considers relationships between documents via a known graph of dependencies, which leaves the scope of this study where we assume that such relational knowledge is not available a priori. 

In the vein of representation learning models \cite{bengio2003} and of the famous \texttt{Word2Vec}, a recent craze for time modeling has elicited various models based on word projections in latent vector spaces such as \cite{eger17} - linear temporal dependencies between word representations -,  \cite{bamler17} - a dynamic \texttt{Skip-Gram} model -, \cite{rudolph17} - a model with exponential probabilistic evolution - or  \cite{yao17} - matrix factorization with temporal alignment. As opposed to \cite{kaban02}, textual tokens are projected in a continuous space rather than on a discrete grid, which enables the use of classical continuous optimization methods. Moreover, contrary to previous approaches based on topic distributions with temporal dependencies, the goal of these works is to learn some semantic representations of words that can be used directly in various neural models.
The temporal dependencies are defined on word representations: each considered time-step is associated with its own vocabulary representation forced to respect various temporal constraints. However, it appears difficult to consider such a kind of approach in a multi-author setting, for which separated representations should be learned both per time-step and also per author. We can note the approach of \cite{rudolph17b} for grouped data that enables a reduction of the number of parameters that have to be learned by sharing context vectors between groups, but whose transposition to a multi-author setting appears difficult (very high number of groups, doubled dependencies, temporal evolution vs connected groups).
Another limitation with this kind of approach is that they do not allow end-to-end learning of language models, and extending them for outputting word probabilistic distributions is usually difficult.

An alternative to these various models is to leverage RNNs for language modeling. A recurrent language model takes a sequence of words of arbitrary size as input and outputs a probability distribution of the next word. Such models are often parameterized by LSTM networks \cite{hochreiter97}. Compared to the skip-gram algorithm that uses a limited context window, recurrent language models operate on sequences of arbitrary length and can capture long-term dependencies. They are nowadays used at the core of an increasing number of tasks, for instance as a feature extractor for text classification \cite{peters18}, as a core building block of unsupervised Neural Machine Translation models \cite{lample2018phrase}, or as a discriminator for Generative Adversarial Models on text \cite{yang18}.

Conditioning language models has already been considered for modeling the context of words in the documents \cite{le14}, but, to the best of our knowledge, not for the extraction of some temporal or structural dynamics in author communities. Rather than defining an individual vectorial representation for each word at every step and for each author, which appears highly too complex to be correctly learned, the idea is to rely on learned author representations modified according to a dynamic function.

\section{Model}
\label{sec:model}

We propose a deep language model that extends classical recurrent methods by incorporating knowledge about the author and the publication time of each document. We learn latent vectors that represent features specific to textual expression modes of the authors. In order to handle temporal drifts, we propose a dynamic model that updates authors' representations through time in the latent space. These latent vectors condition an LSTM language model, allowing it to adapt its own dynamics depending on language bias specific to authors and timesteps.

In \secref{sec:model-task}, we present our notations and task. In \secref{sec:model-model}, we present our dynamical language model based on temporal author representations. And in \secref{sec:model-dynamic}, we describe how we update author representations through time by learning a residual dynamic function in the latent space.

\subsection{Notations and Task}
\label{sec:model-task}
We consider text publications defined over a vocabulary of size $V$. Let $A$ be the set of considered authors with texts published in the time interval $\{1,\dots, T\}$. We formulate the problem as maximizing the likelihood of a textual document $\vx$ knowing its author $a \in A$ and its publication timestep $t \in \{1,\dots, T\}$:
\begin{equation}
    P(\vx | a, t) = \prod\limits_{k=0}^{|\vx|} P(\evx_{k+1} | \vx_{0:k}, a, t),
    \label{eq:ll}
\end{equation}
where $\evx_k$ is the $k$\textsuperscript{th} token of $\vx$, $|\vx|$ is the number of tokens in $\vx$. $\evx_0$ is a start-of-sentence token and $\evx_{|\vx| + 1}$ is an end of sentence token. The notation $\vx_{0:k}$ refers to tokens $\{\evx_{0},\dots, \evx_{k}\}$. Note that author $a$ may have published 0, 1, or several documents at a particular timestep $t$. So, another challenge of our task is to handle gaps in author publication histories, and the uneven distribution of documents among authors, and through time.

\subsection{A Dynamic Language Model}
\label{sec:model-model}

\begin{figure}[t]
\centering
\includegraphics[width=0.8\linewidth]{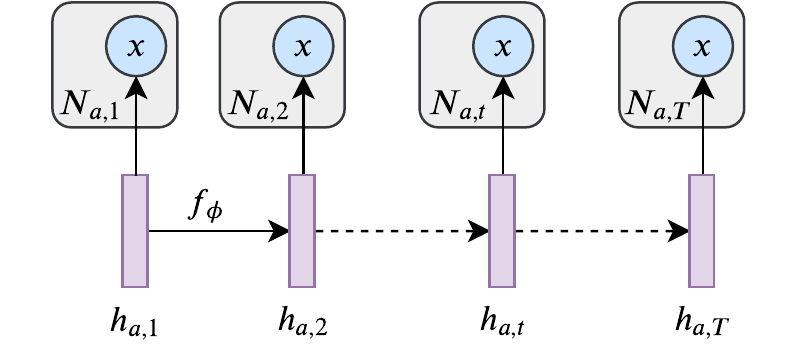}
\caption{High-level view of our proposed dynamic language model for an author $a$. $h_{a,t}$ are the conditioning vectors that evolve through time with a dynamic function $f_{\phi}$. $\vx$ are text publications at different timesteps and $N_{a, t}$ is the number of texts published by author $a$ at timestep $t$. The panels surrounding each variables $\vx$ highlight the fact that several documents ($N_{a,t}$) are modeled conditionally on the same vector $h_{a,t}$.}
\label{fig:model-hl}
\end{figure}

The language modeling task is auto-regressive, as shown in \eqref{eq:ll}, making recurrent neural networks, and particularly LSTMs, the most natural deep learning methods to handle this task. They are currently at the state of the art for language modeling \cite{melis2017state, merity2017regularizing}. We thus choose to construct our method on an LSTM network that we condition to an author $a$ and a timestep $t$ through a latent vector $h_{a,t}$. We now consider that all the information specific to the author $a$ at time $t$ is contained in this vector. The probability of a document $\vx$ written by $a$ at time $t$ for an LSTM with parameters $\theta$ is defined as follows:
\begin{equation}
    P(\vx | a, t) = P_{\theta}(\vx | h_{a, t}) = \prod\limits_{k=0}^{|\vx|} P_{\theta}(\evx_{k+1} | \vx_{0:k}, h_{a, t}).
    \label{eq:ll-theta}
\end{equation}
An overview of our approach is pictured in \figref{fig:model-hl}.

In this setting, we can view the LSTM as a decoder: it takes as input a conditioning vector $h_{a,t}$ and a word history $\evx_{0:k}$, and outputs the next word probability distribution, as formulated in the right-hand side of \eqref{eq:ll-theta}. We experimented with several methods to incorporate $h_{a,t}$ into the LSTM. We found that projecting $h_{a,t}$ into the word embeddings space, and using it in place of the start of sentence token yields the best results. This is consistent with other works \cite{subramanian2018multiple}. The intuition is that it prevents the LSTM from overfitting, compared to other approaches (e.g. concatenating the latent vector at each timestep). Since our experiments are performed on relatively short documents, we did not have problems with the LSTM forgetting the conditioning.

\subsection{Dynamic Author Representation}
\label{sec:model-dynamic}
In this section, we present the dynamic conditioning of the language models, corresponding to the $f_{\phi}$ function depicted in \figref{fig:model-hl}. Depending on the way the condition $h_{a,t}$ is defined for a step $t$ and an author $a$, the model can greatly differ in the dynamics/dependencies it captures.

The general idea of the model is to produce a latent trajectory for each author. A latent trajectory is a sequence of representation vectors $h_{a, t}$ that evolve in time with a function $f_{\phi}$ parametrized by $\phi$. The general formulation is as follow:
\begin{equation*}
    h_{a, t} = f_{\phi}(h_{a, 0}, ..., h_{a, t-1}).
\end{equation*}
The formulation is fairly general, and several architectures can fit $f_{\phi}$.

The challenge in learning the $h_{a, t}$ vectors is twofold. First, they should capture features specific to author $a$ that do not change in time. For instance, in the case of a scientific community, the scientific scope of an author (computer science, physics, biology, etc...) usually do not change through the years. And second, it should capture the variations in authors expression mode and topic evolution through time. The writing style of an author may indeed change through time, and its topics of interests may also change more or less drastically.

\begin{figure}[t]
\centering
\includegraphics[width=0.8\linewidth]{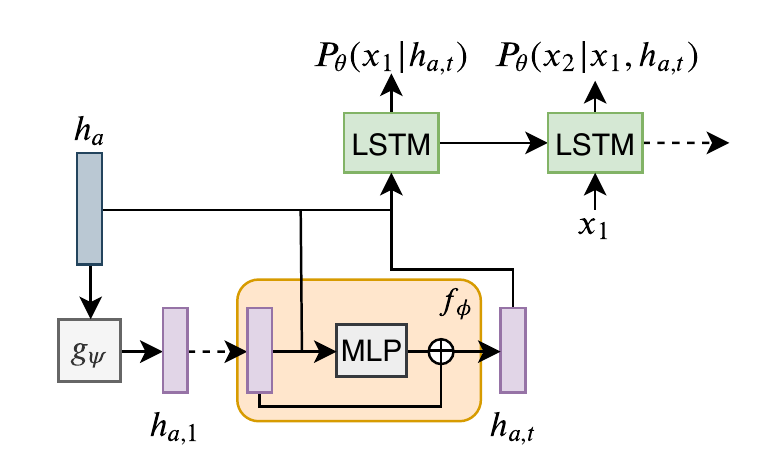}
\caption{Detailed view of the proposed architecture. The initialization function $g_{\psi}$ uses the static representation of author $a$, $h_{a}$, to produce the first latent vector $h_{a,1}$. The residual function $f_{\phi}$ is then recursively applied in order to produce $h_{a, t}$, which is used by the LSTM decoder to model a text sequence $\vx$ written by $a$ at $t$.}
\label{fig:model-det}
\end{figure}

To facilitate the learning of static features, we learn a latent vector $h_a$ per author. These vectors are constant through time and used in various ways in our model. It allows the dynamic function to focus only on variations across timesteps, as described below. 

We use a residual architecture for our dynamic function. We chose a Markovian transition function, which only considers the previous representation $h_{a,t-1}$, for the induction of $h_{a,t}$. It appears as a good trade-off between robustness and flexibility. More powerful sequential models, such as RNNs that maintain a memory of the past states, would be prone to overfitting. Indeed, the number of authors and timesteps is usually small compared to the number of documents in the collections, and lots of author-timestep pairs are missing. Having a residual function in our dynamics allows us to learn smooth trajectories, as the magnitude and direction of the residue can be constrained easily by regularizing $\phi$ with an L2 norm. This dynamic function writes as follows:
\begin{equation*}
    h_{a, t} = h_{a, t-1} + f_{\phi}(h_{a, t-1}, h_a).
    \label{eq:res-dyn}
\end{equation*}

In this case, $f_{\phi}$ is a Multi-Layer Perceptron (MLP) with ReLU activations. In addition to the previous state, we also give the static representation $h_{a}$ to the MLP in order to encourage different dynamics among authors. Without it, two representations at the same position in the latent space would have the same next state, and hence the same following dynamics. We also use $h_{a}$ to compute the initial vector $h_{a, 1}$ through a specific MLP, $g_{\psi}$.

Finally, $h_{a, t}$ vectors are concatenated to the static author representations $h_a$ to form the conditioning vectors that are fed to the LSTM decoder. Since the decoder is also fed sequentially with a word context $\evx_{1:k}$, an encoder is not needed. The decoder is thus able to capture general language structure, like grammar, and use the conditioning vectors to adapt its internal dynamic to a specific author at a specific timestep. A detailed view of the described architecture is pictured of \figref{fig:model-det}.

\section{Experimental Setup}
\label{sec:setup}

\begin{table*}[t]
    \caption{Perplexity on the Semantic Scholar corpus}
    \centering
    \label{tab:s2-ppl}
    \begin{tabular}{|l|rr|rr|rr|}
        \hline
        \multicolumn{1}{|c|}{} & \multicolumn{2}{c|}{\textbf{Modeling}} & \multicolumn{2}{c|}{\textbf{Imputation}} & \multicolumn{2}{c|}{\textbf{Prediction}} \\
        \hline
        \textbf{Models} & \textit{micro} $\pm$ \textit{stdv} & \textit{macro} $\pm$ \textit{stdv} & \textit{micro} $\pm$ \textit{stdv} & \textit{macro} $\pm$ \textit{stdv} & \textit{micro} $\pm$ \textit{stdv} & \textit{macro} $\pm$ \textit{stdv}\\
        \hline
        LSTM & $53.8 \pm 0.07$ & $65.0 \pm 0.35$ & $57.4 \pm 0.07$ & $71.5 \pm 0.21$ & $80.7 \pm 0.16$ & $83.0 \pm 0.52$ \\
        LSTM-A & $48.0 \pm 0.11$ & $56.8 \pm 0.67$ & $52.7 \pm 0.08$ & $63.9 \pm 0.45$ & $77.2 \pm 0.26$ & $\mathbf{77.8} \pm 0.88$ \\
        LSTM-iAT & $54.3 \pm 0.08$ & $68.2 \pm 0.80$ & $61.3 \pm 2.82$ & $77.1 \pm 4.69$ & $83.7 \pm 0.17$ & $88.0 \pm 0.86$ \\
        LSTM-AT & $47.7 \pm 0.09$ & $55.4 \pm 0.22$ & $52.3 \pm 0.08$ & $62.9 \pm 0.31$ & $77.2 \pm 0.13$ & $\mathbf{77.3} \pm 1.31$  \\
        Ours & $\mathbf{46.7} \pm 0.09$ & $\mathbf{53.3} \pm 0.22$ & $\mathbf{51.2} \pm 0.09$ & $\mathbf{60.2} \pm 0.20$ & $\mathbf{74.3} \pm 0.23$ & $\mathbf{77.5} \pm 1.22$\\
        \hline
    \end{tabular}
\end{table*}

\begin{table*}[t]
    \caption{Perplexity on the New York Times corpus}
    \centering
    \label{tab:nyt-ppl}
    \begin{tabular}{|l|rr|rr|rr|}
        \hline
        \multicolumn{1}{|c|}{} & \multicolumn{2}{c|}{\textbf{Modeling}} & \multicolumn{2}{c|}{\textbf{Imputation}} & \multicolumn{2}{c|}{\textbf{Prediction}} \\
        \hline
        \textbf{Models} & \textit{micro} $\pm$ \textit{stdv} & \textit{macro} $\pm$ \textit{stdv} & \textit{micro} $\pm$ \textit{stdv} & \textit{macro} $\pm$ \textit{stdv} & \textit{micro} $\pm$ \textit{stdv} & \textit{macro} $\pm$ \textit{stdv}\\
        \hline
        LSTM & $112.4 \pm 0.22$ & $112.9 \pm 0.23$ & $108.8 \pm 0.11$ & $109.4 \pm 0.21$ & $114.5 \pm 0.17$ & $110.1 \pm 0.17$ \\
        LSTM-A & $100.1 \pm 0.22$ & $100.7 \pm 0.21$ & $100.7 \pm 0.13$ & $101.3 \pm 0.20$ & $113.1 \pm 0.30$ & $108.25 \pm 0.34$ \\
        LSTM-iAT & $108.9 \pm 0.34$ & $110.0 \pm 0.38$ & $135.8 \pm 0.59$ & $136.6 \pm 0.56$ & $121.0 \pm 0.57$ & $115.9 \pm 0.52$ \\
        LSTM-AT & $\mathbf{97.3} \pm 0.10$ & $\mathbf{97.9} \pm 0.09$ & $98.9 \pm 0.20$ & $99.5 \pm 0.23$ & $113.1 \pm 0.19$ & $108.3 \pm 0.21$ \\
        Ours & $\mathbf{97.1} \pm 0.14$ & $\mathbf{97.7} \pm 0.14$ & $\mathbf{98.2} \pm 0.25$ & $\mathbf{98.7} \pm 0.24$ & $\mathbf{110.8} \pm 0.38$ & $\mathbf{106.5} \pm 0.34$ \\
        \hline
    \end{tabular}
\end{table*}

We evaluate the proposed model together with several temporal and non-temporal baselines described in \secref{sec:setup-baselines}. We propose to evaluate the models in three temporal settings: modeling, imputation and prediction presented in \secref{sec:setup-tasks}, on two temporal corpora described in \secref{sec:setup-dataset}.

\subsection{Model and Baselines}
\label{sec:setup-baselines}
We compare the following models:
\begin{itemize}
    \item \textbf{LSTM}: a classical LSTM decoder (no conditioning on the publication time or the authors). We use this model to assess the gain in performances of our model and other baselines
    \item \textbf{LSTM-A}: an LSTM decoder conditioned on authors embeddings. Only $h_a$ is given as the start token of the LSTM decoder. This baseline allows us to assess the performances of our temporal component.
    \item \textbf{LSTM-iAT}: an LSTM decoder conditioned on authors and time with vectors $h_{a,t}$ that are free parameters to learn (no dynamics and no constraints on successive vectors). It is the most naive way to condition a language model on authors and time.
    \item \textbf{LSTM-AT}: similar to LSTM-iAT, but where an L2 regularization between consecutive vectors is applied during learning in order to structure the embedding space. It is a robust baseline, but without a dynamical module to predict representations.
    \item \textbf{Ours}\footnote{code available at \url{https://github.com/edouardelasalles/dar}}: the model described in \secref{sec:model}.
\end{itemize}

\subsection{Evaluation and Tasks}
\label{sec:setup-tasks}

\begin{figure}[t]
\centering
\includegraphics[width=0.7\linewidth]{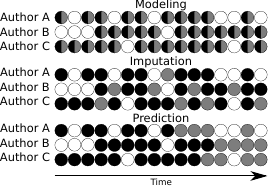}
\caption{Illustration of our three tasks. A, B, and C are three authors, and each column a timestep. A circle represents a set of documents published by a given author at a given timestep. Black circles are training data, grey circles test data, and white circle missing data. Validation data were omitted for simplification purposes.}
\label{fig:tasks}
\end{figure}

We quantitatively evaluate our model and baselines for the language modeling task. We compare models based on their token perplexity. Results reported in \secref{sec:xp} were obtained on held-out test sets. Model and hyperparameters selection were performed with a separate validation set. The split proportions between training, validation, and testing sets are always approximately 70\% / 10\% / 20\%. Each experiment was run 5 times with different seeds, and the reported results are the mean and standard deviation across these 5 runs.

We compare our model and the baselines in 3 temporal settings, which are depicted in figure \ref{fig:tasks}. Each evaluation setting corresponds to a distribution of the train / val / test splits across timesteps. A setting can be seen as a temporal task, and help us analyze different behaviors. The three temporal tasks are:
\begin{itemize}
    \item \textbf{Modeling}: documents published at every timestep are visible by the model during training. The train / val / test splits are sampled randomly, with the constraint of keeping the same distribution of authors across all splits. It is the easiest setting, as there is a least one document in the train set for each author-timestep couple in the test set.
    \item \textbf{Imputation}: we hide all documents published at randomly chosen timesteps for each author in the train set. For each author, different timesteps are kept. This means that all documents written by author $a$ at time $t$ are either in the train, validation, or test set. This task allows us to assess the smoothness of the learned representations.
    \item \textbf{Prediction}: only the first documents (in chronological order) for each author are visible by the model during training. Since every author has not the same publication rate, the train set stops at different steps for different authors, as depicted in figure \ref{fig:tasks}. It is the most difficult setting, as the models must manage to extrapolate the author's representations.
\end{itemize}

LSTM-iAT and LSTM-AT baselines are not equipped to predict latent representations. So, when evaluating documents published by author $a$ at timesteps $t$ where no document was visible during training, we use the latent representation $h_{a, t'}$, with $t' < t$ the most recent timesteps where documents were present during training. For our method, we use the dynamic function $f_{\phi}$ to predict the representation $h_{a, t}$.

\subsection{Datasets}
\label{sec:setup-dataset}
We evaluate the proposed model on two different corpora presented below:
\begin{itemize}
\item [-] The \textbf{Semantic Scholar} \cite{ammar2018} corpus (S2) is composed of titles from scientific papers published in machine learning conferences and journals from 1985 to 2017, split by year ($33$ timesteps). We lower-cased the texts and used the same WordPiece model as in \cite{devlin18} to tokenize the corpus, which has around 30K tokens. The corpus is composed of 45K titles, representing a total of 800K tokens with 1000 authors. The number of titles is not uniformly distributed, and grows quasi-exponentially with time: the year 1985 contains around 100 documents while the year 2017 has around 5K.
\item [-] The \textbf{New York Times} \cite{yao18} corpus (NYT) is composed of headlines from the New York Times newspaper spanning from 1990 to 2015, also split by years ($26$ timesteps). We also lower-cased the texts, but we use the NLTK \cite{loper02nltk} word tokenizer, and replaced every number with a special \texttt{N} token. Words appearing less than 5 times in the training set were discarded, giving a vocabulary of around 6K tokens\footnote{We did not use WordPiece in the NYT corpus since we noticed in experiments that it led to dramatic overfitting.}. The corpus contains 40K documents, 470K tokens, and 500 authors. In this corpus, the documents are evenly distributed in time.
\end{itemize}

\subsection{Architectural and Optimization Details}
\label{setup-xp}
For both corpora, the LSTM decoder is two-layer AWD-LSTM \cite{merity2017regularizing} with hidden units and word embeddings of size 400. We use weight dropout, variational dropout, and embeddings weight-tying. We use the Adam optimizer with mini-batches of size 64, a learning rate of 0.003, and default parameters. Learning rate is constant for 50K iterations for S2, and 30K for NYT, and then decreased linearly for 20K iterations for S2 and 5K for NYT. Models were trained on a TITAN Xp Pascal GPU. Models on S2 converge in about 1 hour, and 30 minutes on NYT.

Hyperparameters were tuned on a dedicated validation set. The dropouts were tuned for the LSTM baseline on the modeling task and kept constant across all models and tasks for a given corpus. The weight decay and hyperparameters specific to each model were tuned independently by grid search.

\section{Results}
\label{sec:xp}
We present the language modeling results in \secref{sec:xp-quanti}. In sections \ref{sec:xp-visu-latent}, \ref{sec:xp-lat-s2}, and \ref{sec:xp-lat-nyt} we present analyses of the learned representations. In addition, we present text samples generated by our model in \secref{sec:xp-gen}.

\subsection{Temporal Language Modeling}
\label{sec:xp-quanti}

\begin{figure*}[t]
     \centering
     \begin{subfigure}[b]{0.32\textwidth}
         \centering
         \caption{S2 - Modeling}
         \vspace{-5pt}
         \includegraphics[width=\linewidth]{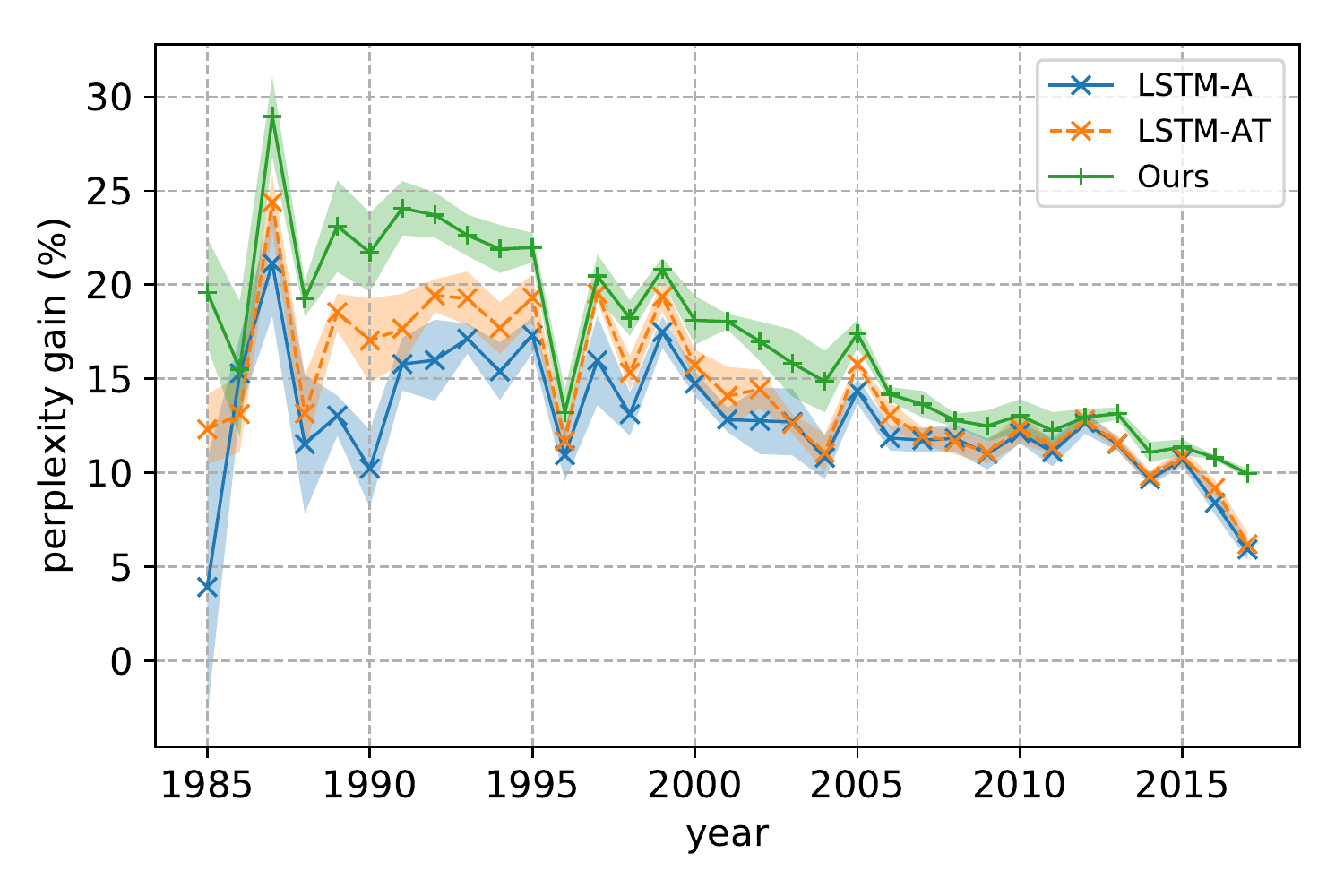}
         \label{fig:gain-s2-modeling}
     \end{subfigure}
     \hfill
     \begin{subfigure}[b]{0.32\textwidth}
         \centering
         \caption{S2 - Imputation}
         \vspace{-5pt}
         \includegraphics[width=\linewidth]{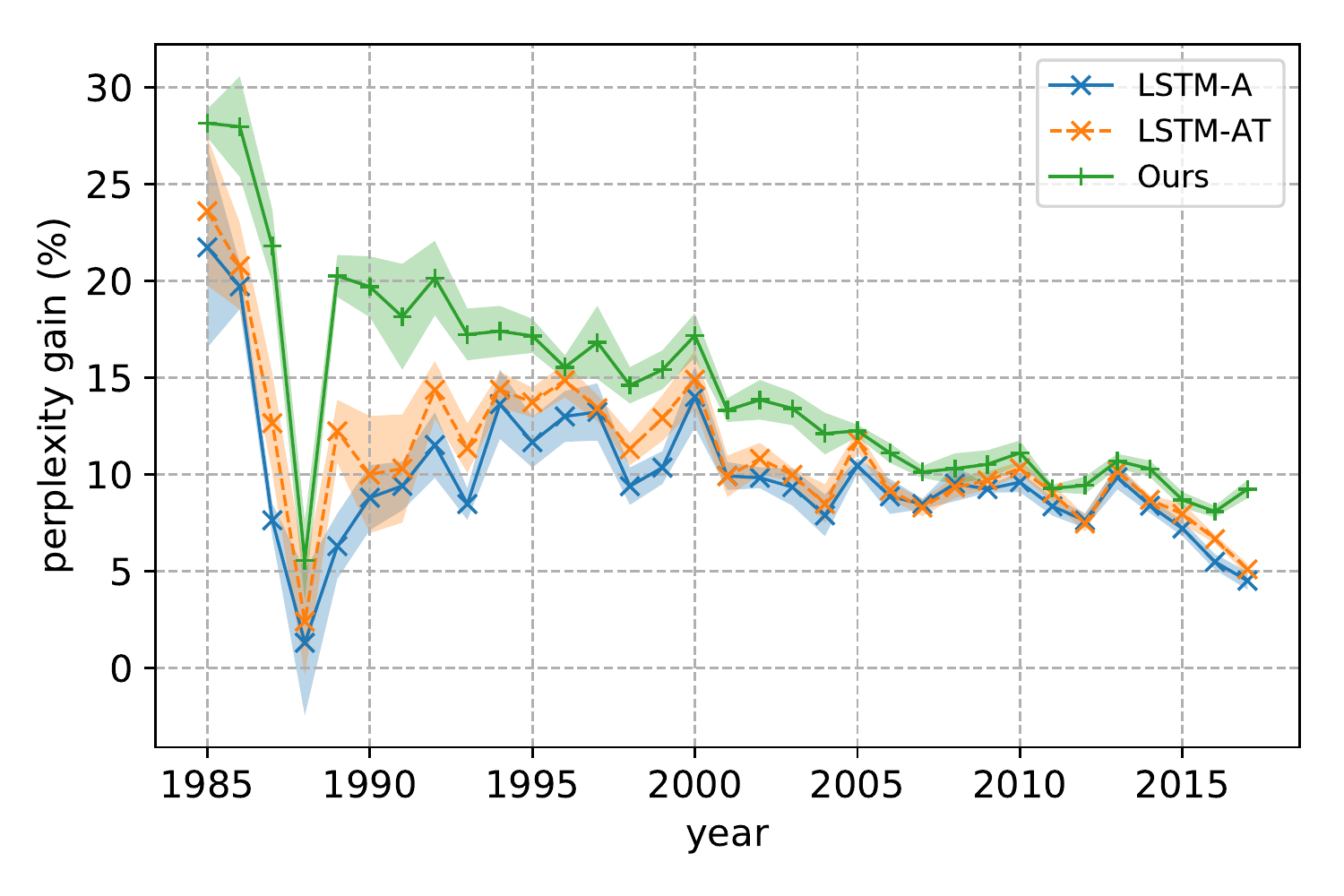}
         \label{fig:gain-s2-imputation}
     \end{subfigure}
     \hfill
     \begin{subfigure}[b]{0.32\textwidth}
         \centering
         \caption{S2 - Prediction}
         \vspace{-5pt}
         \includegraphics[width=\linewidth]{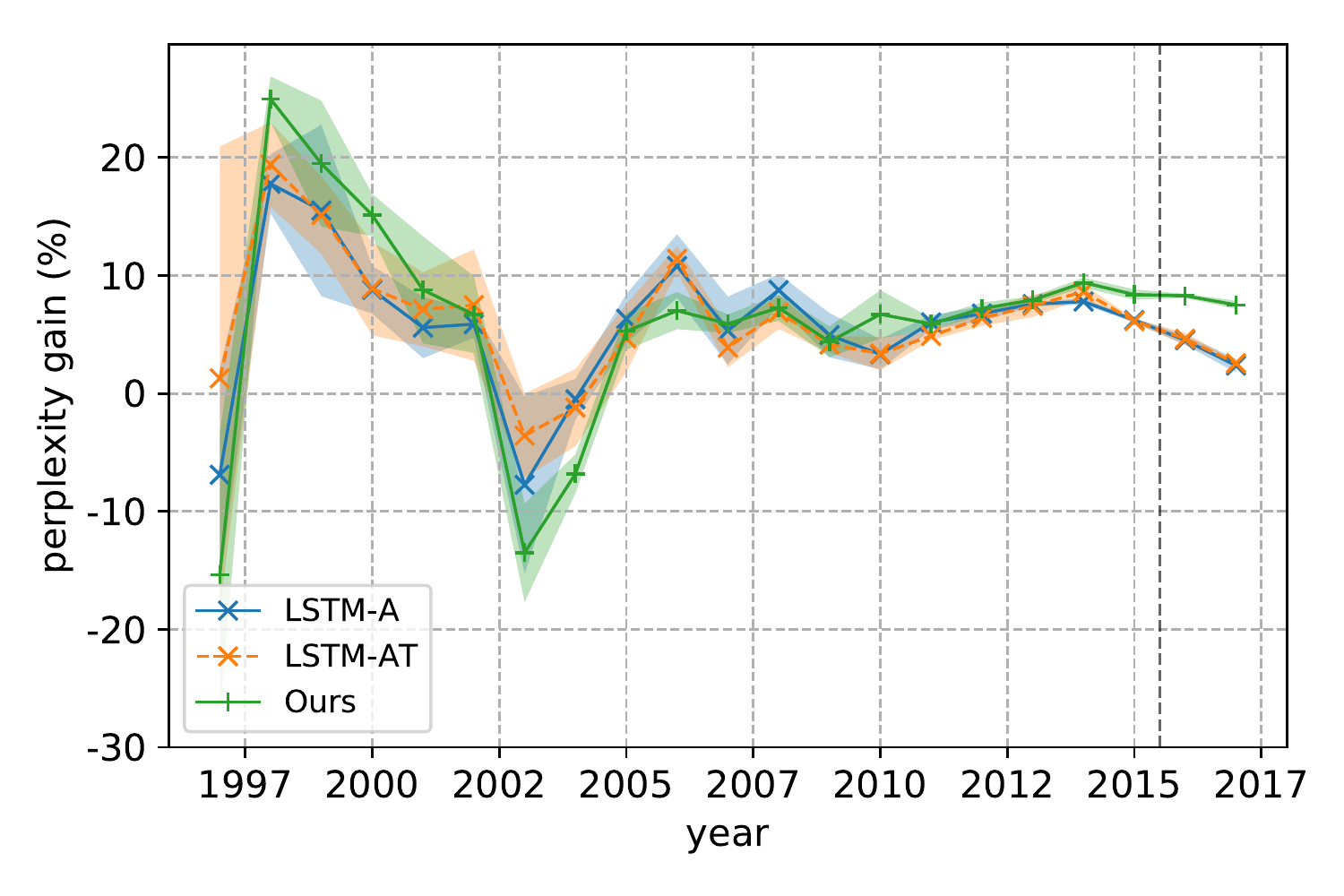}
         \label{fig:gain-s2-pred}
     \end{subfigure}
     \begin{subfigure}[b]{0.32\textwidth}
         \centering
         \caption{NYT - Modeling}
         \vspace{-5pt}
         \includegraphics[width=\linewidth]{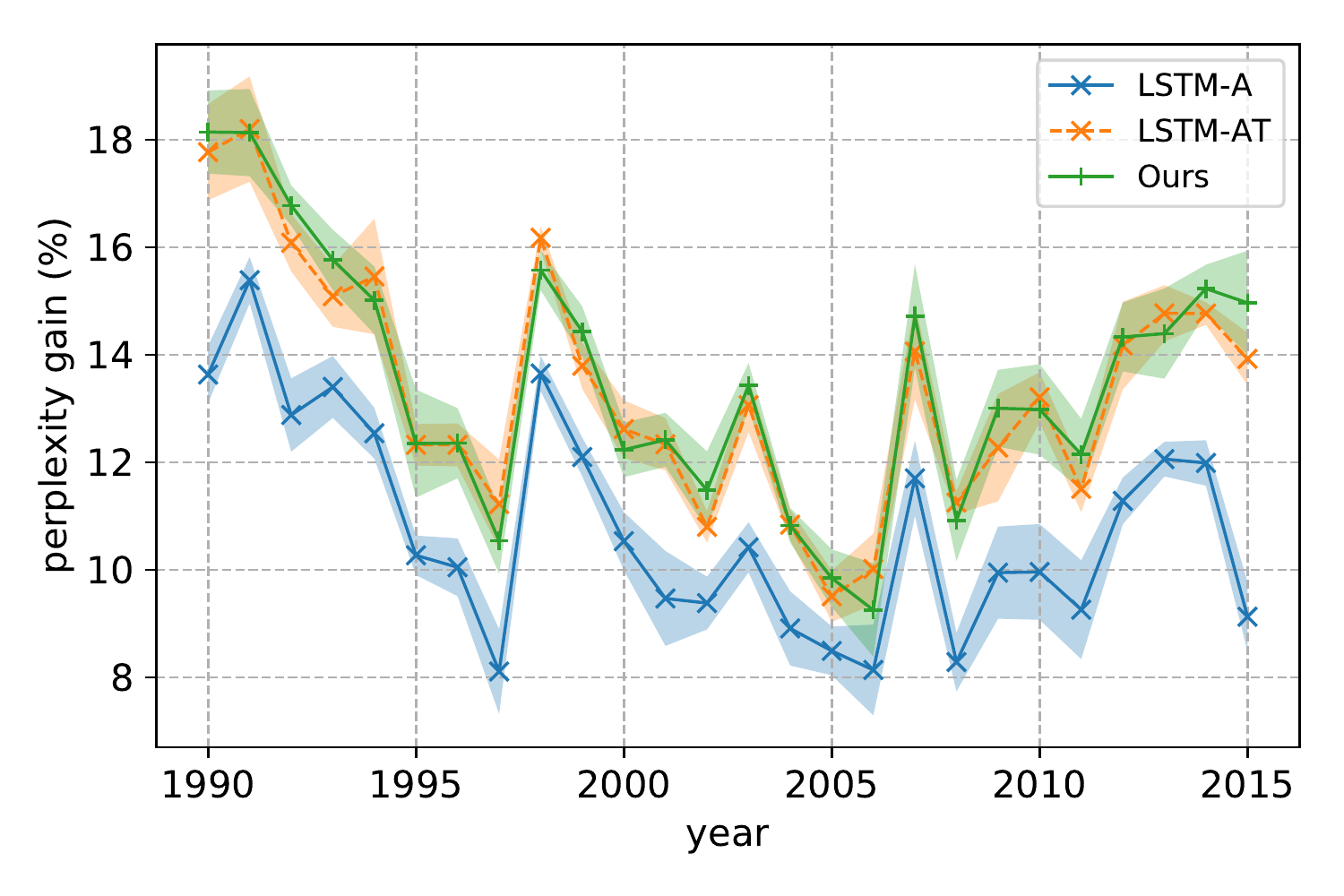}
         \label{fig:gain-nyt-modeling}
     \end{subfigure}
     \hfill
     \begin{subfigure}[b]{0.32\textwidth}
         \centering
         \caption{NYT - Imputation}
         \vspace{-5pt}
         \includegraphics[width=\linewidth]{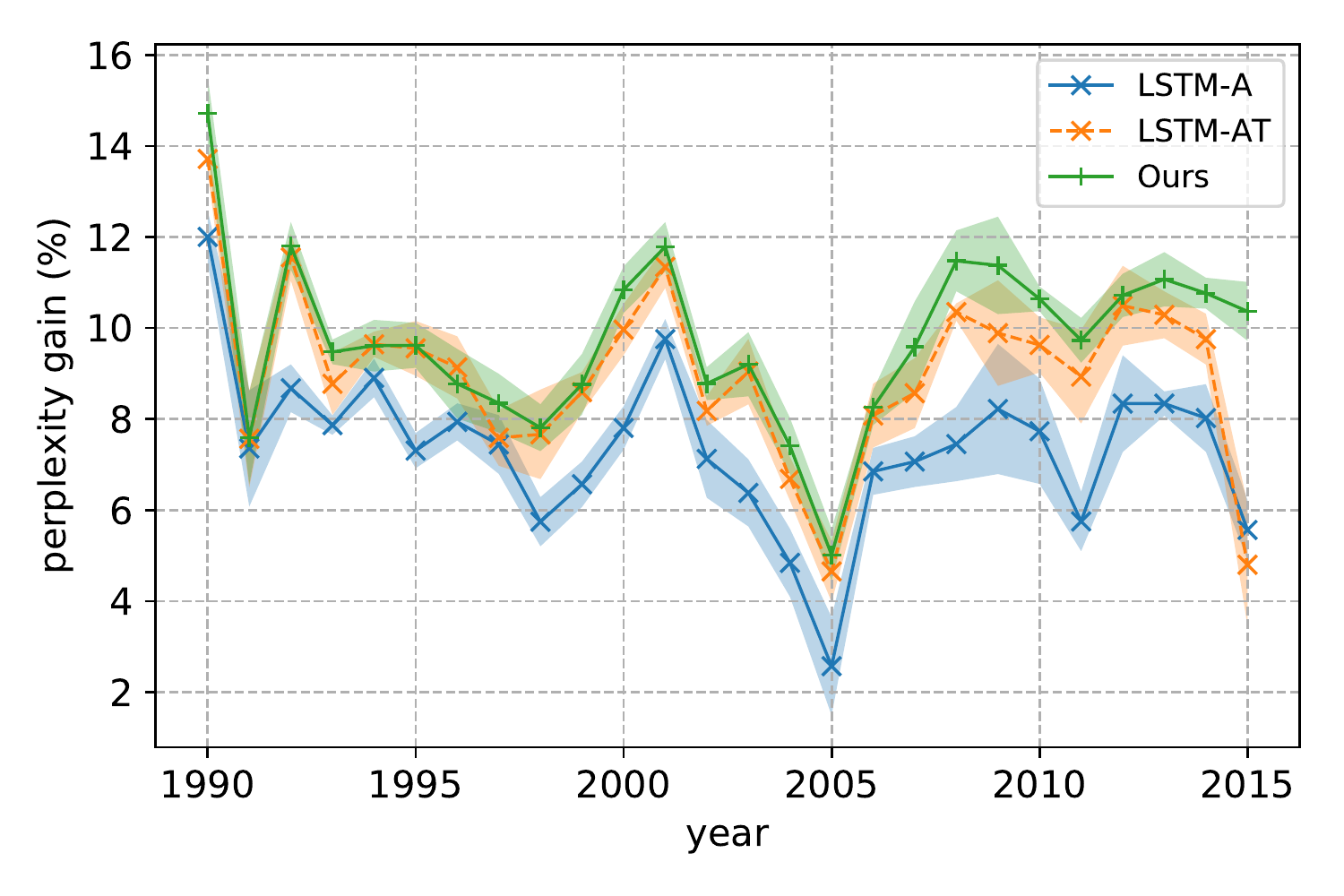}
         \label{fig:gain-nyt-imputation}
     \end{subfigure}
     \hfill
     \begin{subfigure}[b]{0.32\textwidth}
         \centering
         \caption{NYT - Prediction}
         \vspace{-5pt}
         \includegraphics[width=\linewidth]{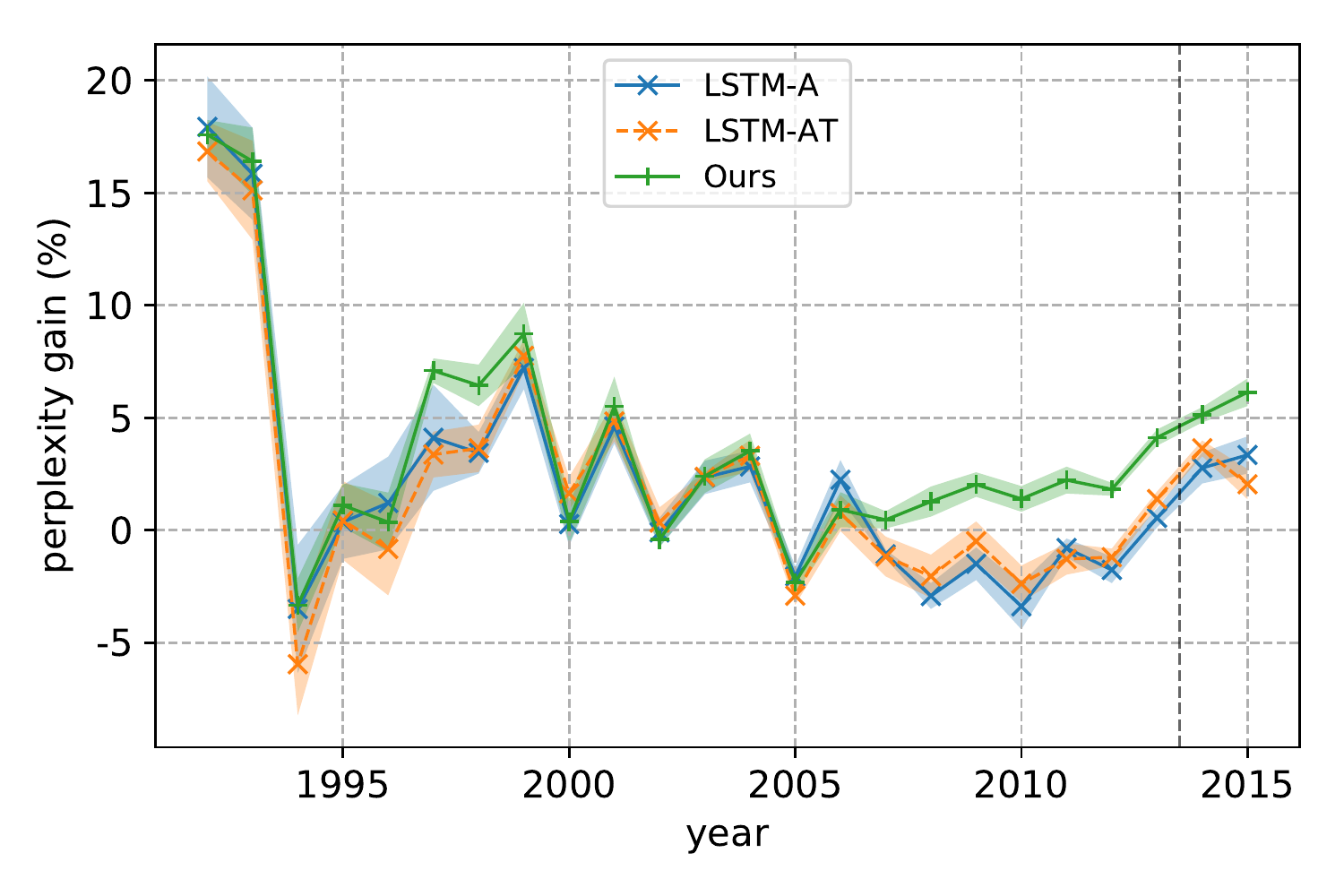}
         \label{fig:gain-nyt-pred}
     \end{subfigure}
    \caption{Perplexity gain w.r.t. the LSTM baseline through time for the Semantic Scholar (top row) and New-York Times (bottom row) corpora (higher is better). The LSTM-iAT baseline is not displayed because it is often significantly worse than the vanilla LSTM, as shown in tables \ref{tab:s2-ppl} and \ref{tab:nyt-ppl}. The black vertical line on the predictions plots represents the point in time from which no documents were seen in the training sets.}
    \label{fig:gain}
\end{figure*}

For each corpus and each task, we show the micro perplexity and the macro temporal perplexity. The micro perplexity is the global token-level perplexity computed indifferently across timesteps. It is the classical language modeling metric that we use to primarily compare models performances. We also provide the macro perplexity, which is the token-level perplexity computed on each timestep separately and then averaged. Since this metric puts the same weight on each timestep, it is possible to see if a model performs consistently across timesteps, even when documents are not evenly distributed in time.

Table \ref{tab:s2-ppl} shows the results on the S2 corpus, and table \ref{tab:nyt-ppl} on NYT. On all tasks and both corpora, our method is significantly better than all the baselines, in micro and macro perplexity. As expected, taking into account authors in a language model improves its performances. But incorporating time into the model is not trivial. LSTM-iAT has consistently worse performances than the vanilla LSTM, except on NYT modeling. Indeed, this baseline tends to overfit, as it has no temporal regularization. In that case, each vector $h_{a,t}$ allows the model to over-specialize itself on texts from the corresponding author $a$ and time $t$.

On S2, LSTM-AT, the temporally regularized version of LSTM-iAT, beats LSTM-A by a small margin (0.2 to 0.4 perplexity points), while our model consistently beats it by 1 to 3 perplexity points, indicating that our dynamic function is more efficient at regularizing the latent presentation on this corpus. On NYT, our method has performances similar to LSTM-AT on the modeling tasks and gains 0.7 points on the imputation task. On the more challenging prediction task, on both S2 and NYT, our model beats LSTM-AT with the greatest perplexity gain across all tasks. We also notice that on this task, LSTM-A and LSTM-AT have the same performances on both corpora. This indicates that our dynamic module is able to accurately predict future states, even at unseen timesteps.
To analyze more specifically the results through time, we show in \figref{fig:gain} the gain in perplexity over the vanilla LSTM through time. For  modeling and imputation on S2 (\figref{fig:gain-s2-modeling} and \ref{fig:gain-s2-imputation}), we can see that our method has the higher gain on every timestep. The gain is more important on the first timesteps, that contains far fewer documents than the last ones. It shows that there is actually a drift in the token distribution and that our model is able to capture it better than a more naive approach. For the same tasks on NYT, we see that LSTM-AT results and ours are similar across timesteps, except for the last ones, where our model maintains the same level perplexity gain while LSTM-AT tends to fall.

For the prediction task (\figref{fig:gain-s2-pred} and \ref{fig:gain-nyt-pred}), we observe similar performances for all models on both corpora. It can be explained by the low number of documents for S2, and the difficulty of the task. On the lasts timesteps however, our model shows a clear gain over the baselines. On S2, the training set contains no documents published at the 2 last timesteps, which is symbolized by the black vertical line in the figure. The low variance and the significant performance gain of our model on these two timesteps indicate that the dynamic module of our model is able to extrapolate at unseen timesteps. On NYT, our model has better results on the last half timesteps. The poor results of LSTM-AT on this task is due in part to the fact that we need to strongly regularize its representation so that it doesn't overfit, and that it does not have a dynamic component, like our model.

\begin{table}
    \caption{Ablation study of the dynamic function $f_{\phi}$.\\ Results are in micro perplexity.}
    \centering
    \label{tab:ab}
    \begin{tabular}{|l|r|r|}
        \hline
        \multicolumn{1}{|c|}{} & \multicolumn{1}{c|}{\textbf{S2}} & \multicolumn{1}{c|}{\textbf{NYT}} \\
        \hline
        ResNet & $47.8 \pm 0.23$ & $100.0 \pm 0.15$ \\
        + AdaDyn & $48.0 \pm 0.45$ & $97.9 \pm 0.26$ \\
        + StatCond & $46.9 \pm 0.13$ & $97.3 \pm 0.16$ \\
        + AdaDyn + StatCond (Ours) & $46.7 \pm 0.09$ & $97.1 \pm 0.14$ \\
        \hline
    \end{tabular}
\end{table}

In \secref{sec:model-dynamic}, we proposed to use a static representation vector $h_a$ in our model. It is used as an additional input to the dynamic function $f_{\phi}$, to adapt its dynamic (AdaDyn) to each author. And as static conditioning (StatCond) of the LSTM decoder, in order to relax $h_{a,t}$ and allowing to focus more on temporal variations. To assess the contribution of these two features into the final results, we performed an ablation study where we removed each feature individually, and altogether. The results are shown in table \ref{tab:ab}. For both corpora, it is the addition of the two features together that yields the best results. StatCond always increases the performances significantly, as it helps the dynamic module to focus only on drifts. On S2, contrary to NYT, the AdaDyn alone does not improve performances of the base ResNet. It means that on this corpus the network does not need to learn individual dynamics for each author, but only a global drift (Adadyn and/or StatCond being required for learning individual trajectories, as explained in section \ref{sec:model-dynamic}). In the next section, we analyze the learned latent trajectories to confirm this behavior.

\subsection{Latent Trajectories Visualization}
\label{sec:xp-visu-latent}
In order to gain a better understanding of our model behavior, we investigate the temporal author representations learned by our model. All the visualizations in this section were extracted from a model learned on the modeling task.

\begin{figure}[t]
     \centering
     \begin{subfigure}[b]{0.49\linewidth}
         \centering
         \caption{S2 with AdaDyn}
         \vspace{-5pt}
         \includegraphics[width=\linewidth]{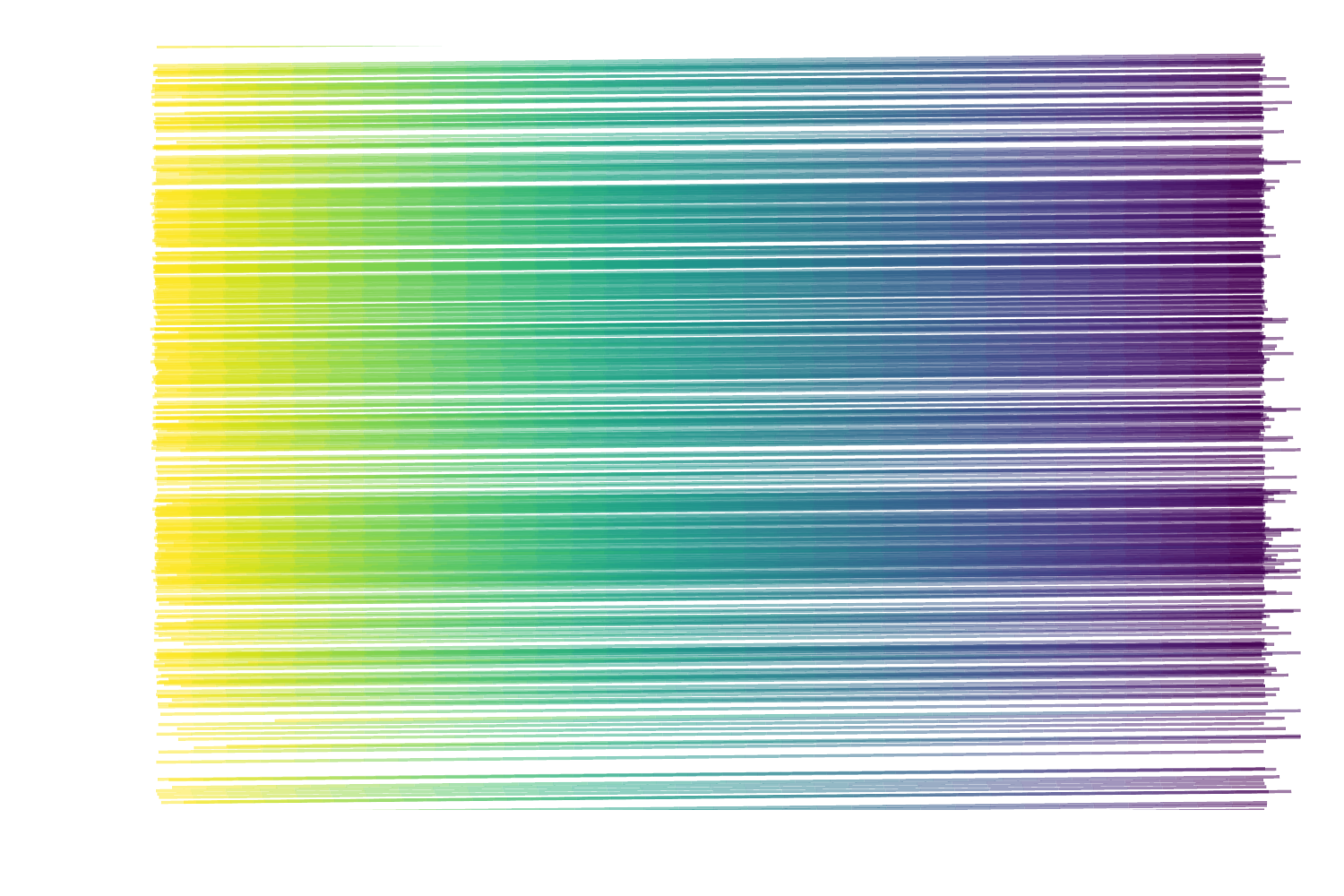}
         \label{fig:latent-s2-sc}
     \end{subfigure}
     \hfill
     \begin{subfigure}[b]{0.49\linewidth}
         \centering
         \caption{S2 without AdaDyn}
         \vspace{-5pt}
         \includegraphics[width=\linewidth]{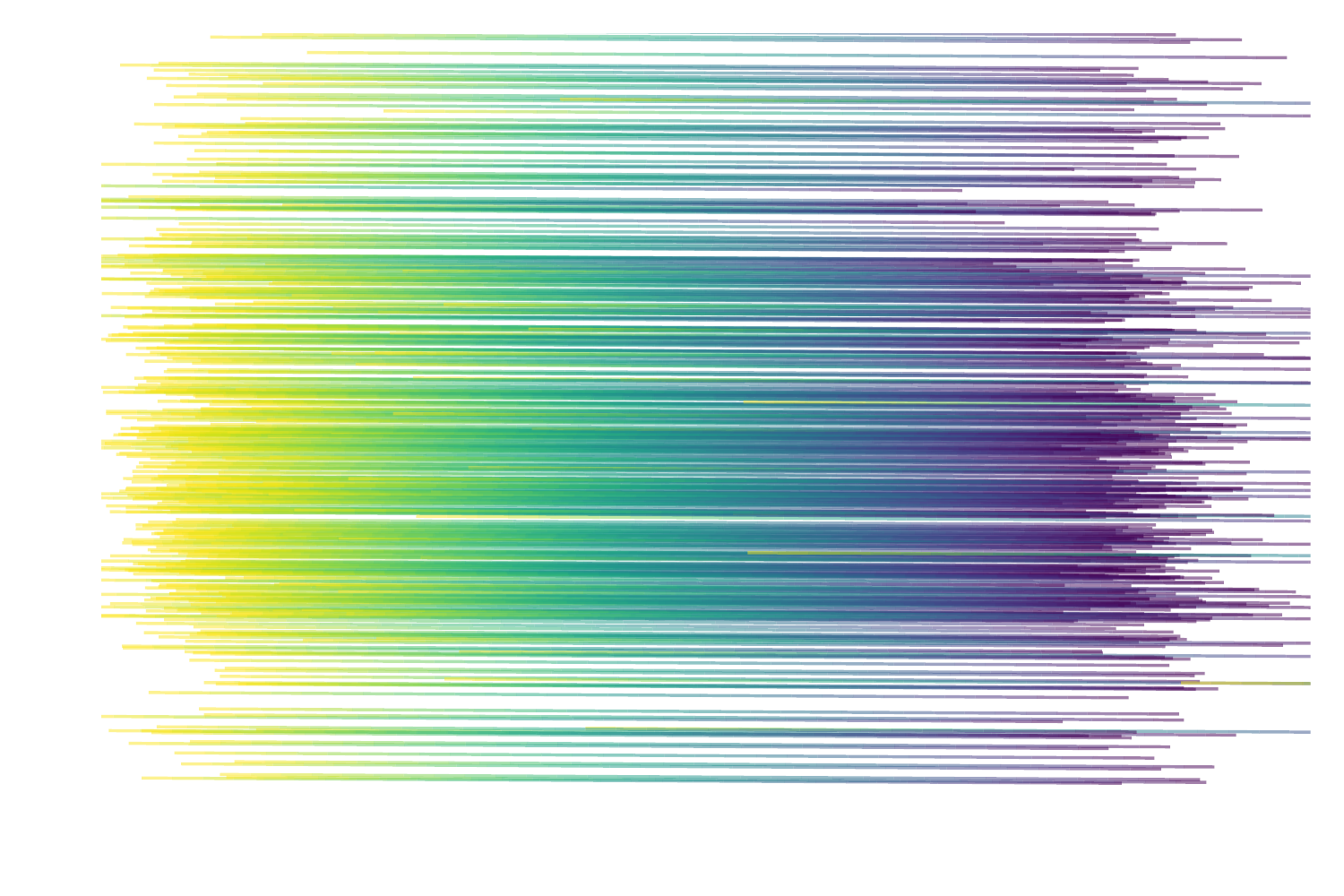}
         \label{fig:latent-s2-nosc}
     \end{subfigure}
     \hfill
     \begin{subfigure}[b]{0.49\linewidth}
         \centering
         \caption{NYT with AdaDyn}
         \vspace{-5pt}
         \includegraphics[width=\linewidth]{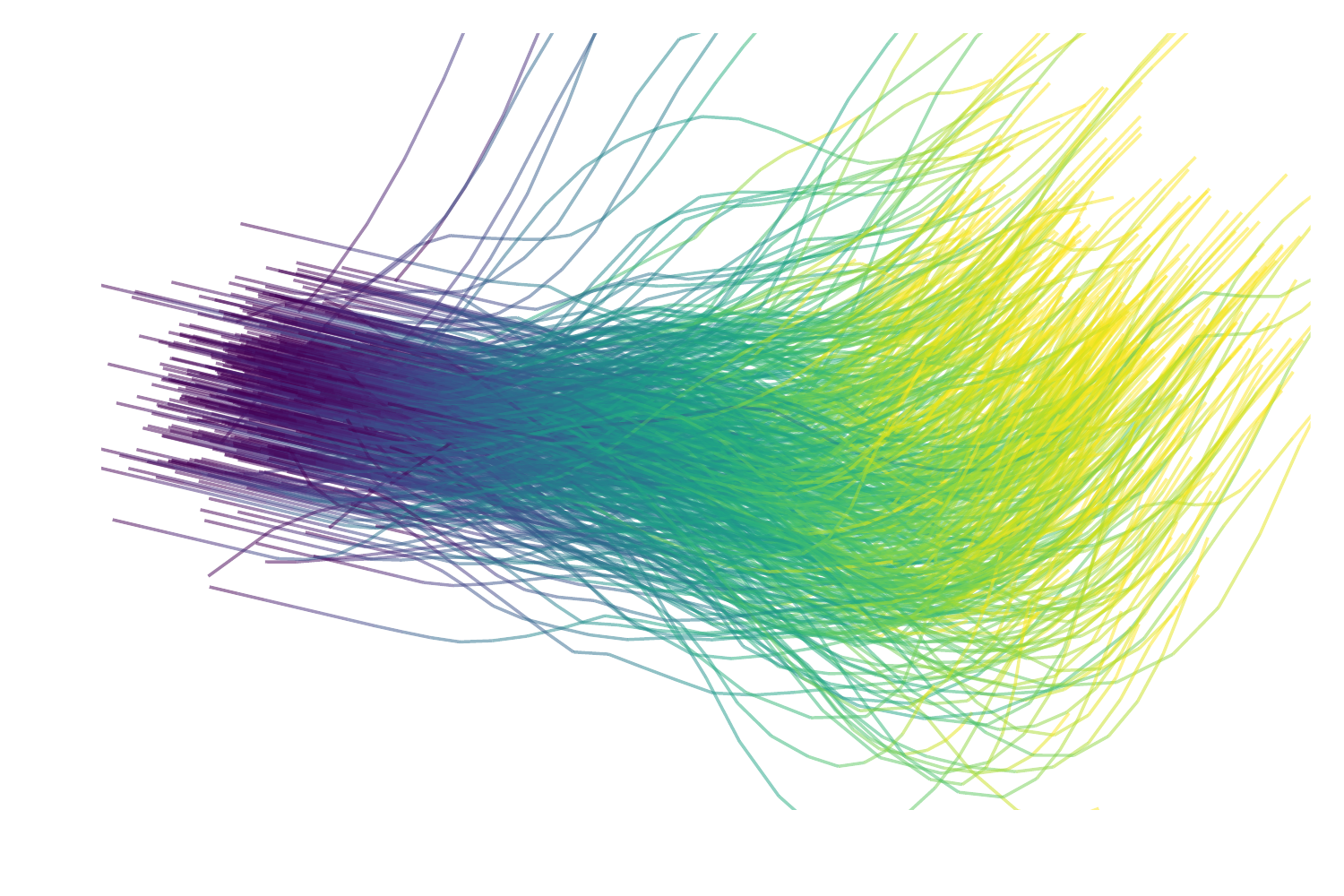}
         \label{fig:latent-nyt-sc}
     \end{subfigure}
     \hfill
     \begin{subfigure}[b]{0.49\linewidth}
         \centering
         \caption{NYT without AdaDyn}
         \vspace{-5pt}
         \includegraphics[width=\linewidth]{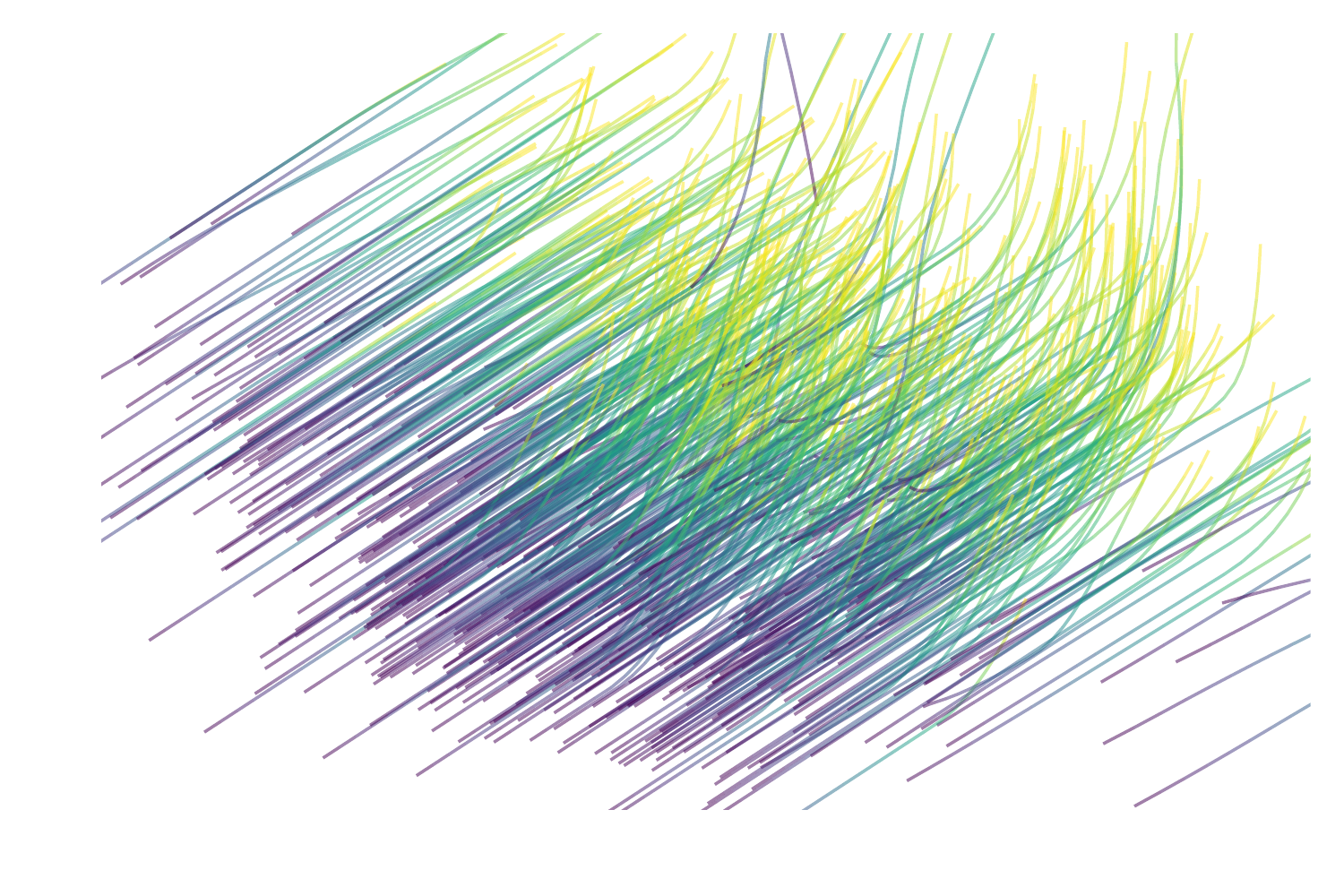}
         \label{fig:latent-nyt-nosc}
     \end{subfigure}
     \caption{PCA of the latent trajectories $h_{a, t}$ for S2 and NYT with and without AdaDyn. Colors represent time: dark at the first timestep to light as the last.}
    \label{fig:latent}
\end{figure}

To visualize the latent trajectories, we performed PCA on the representations and pictured them in \figref{fig:latent}. On NYT, we see that removing the AdaDyn component (\figref{fig:latent-nyt-nosc}) yields parallel trajectories, that all of them drift together in time. On the other hand, with AdaDyn (\figref{fig:latent-nyt-sc}), the dynamic function is free to learn a different dynamic for each author, and we see that the representations drift together in time, but also relatively to each other. On S2 on the other hand, with (\figref{fig:latent-s2-sc}) or without (\figref{fig:latent-s2-nosc}) AdaDyn, the latent trajectories move as one block. It illustrates the results of the ablation study, where we saw that AdaDyn did not improve the results over the ResNet alone on this dataset.

\subsection{Latent Space Analysis: S2}
\label{sec:xp-lat-s2}

In this section, we provide a more detailed analysis of the latent representations learned on the S2 corpus. Since we saw in \secref{sec:xp-visu-latent} that the latent trajectories in S2 do not vary relatively to each other, we focus on here on community-level phenomena.

\begin{figure}[t]
\centering
\includegraphics[width=\linewidth]{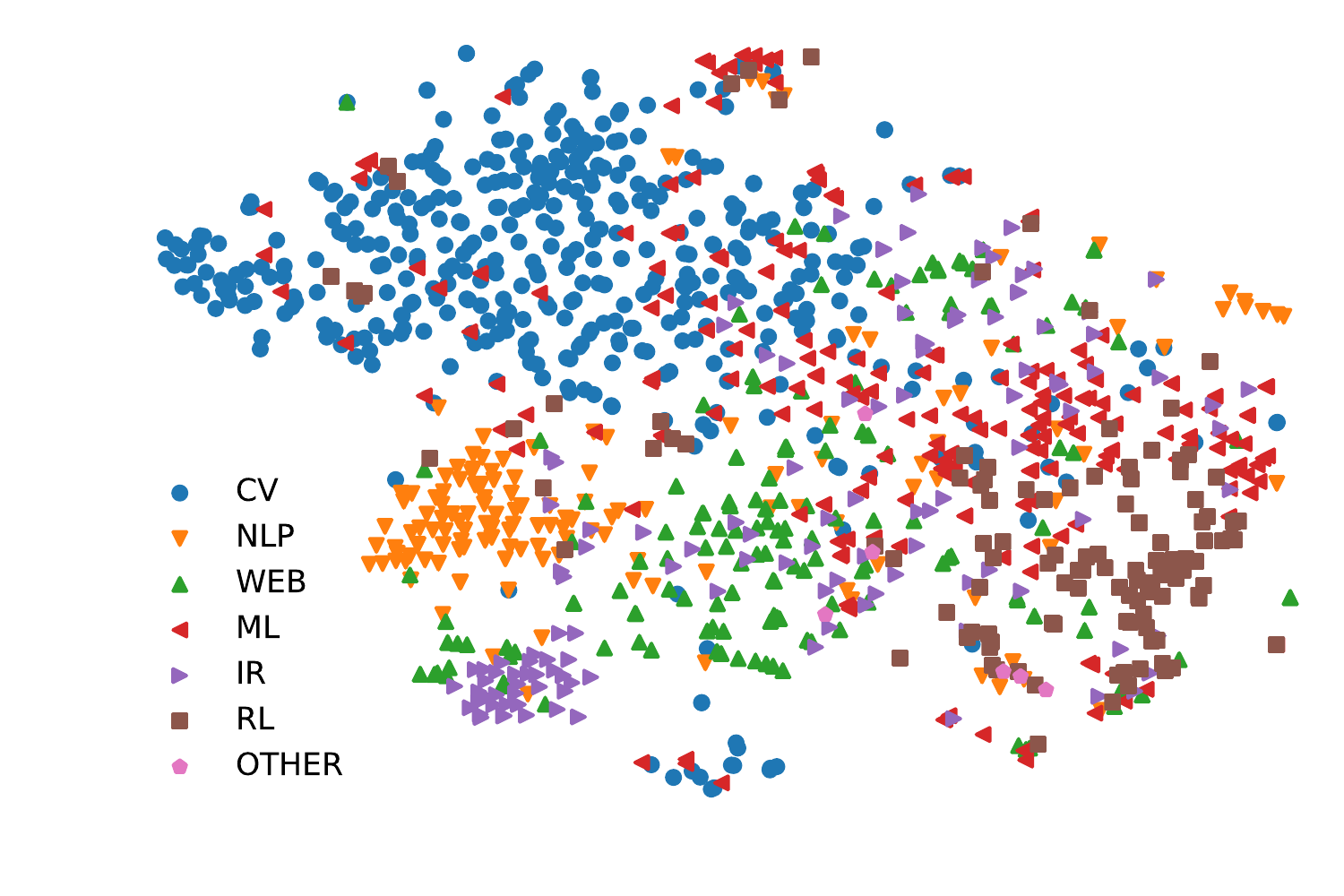}
\caption{t-SNE visualization of the static representations $h_a$ on the S2 corpus.}
\label{fig:s2-he-tsne}
\end{figure}

We begin by plotting on \figref{fig:s2-he-tsne} a t-SNE visualization of the static vectors $h_a$. The labels in this visualization are obtained thanks to key-words associated to each paper in the S2 dataset, that we interpret as topics. We manually clustered the labels into 6 general machine learning categories: Computer Vision (CV), Natural Language Processing (NLP), WEB, Machine Learning (ML), Information Retrieval (IR), and Reinforcement Learning (RL). We also put a category OTHER for authors that do not fit in these categories. We label the authors with the most represented category among their publications. We see on the figure that authors from the CV and NLP communities are distinctly clustered. Next to the NLP cluster, we notice a small IR cluster. Next to these two clusters are several authors from the WEB community. RL authors have their own cluster on the right, though less distinct from the others. And finally, the machine learning authors are spread across all the space, which is expected because the category is very broad since the corpus contains only machine learning papers. It indicates that our static vectors capture semantic information about authors.

\begin{figure}[t]
\centering
\includegraphics[width=0.8\linewidth]{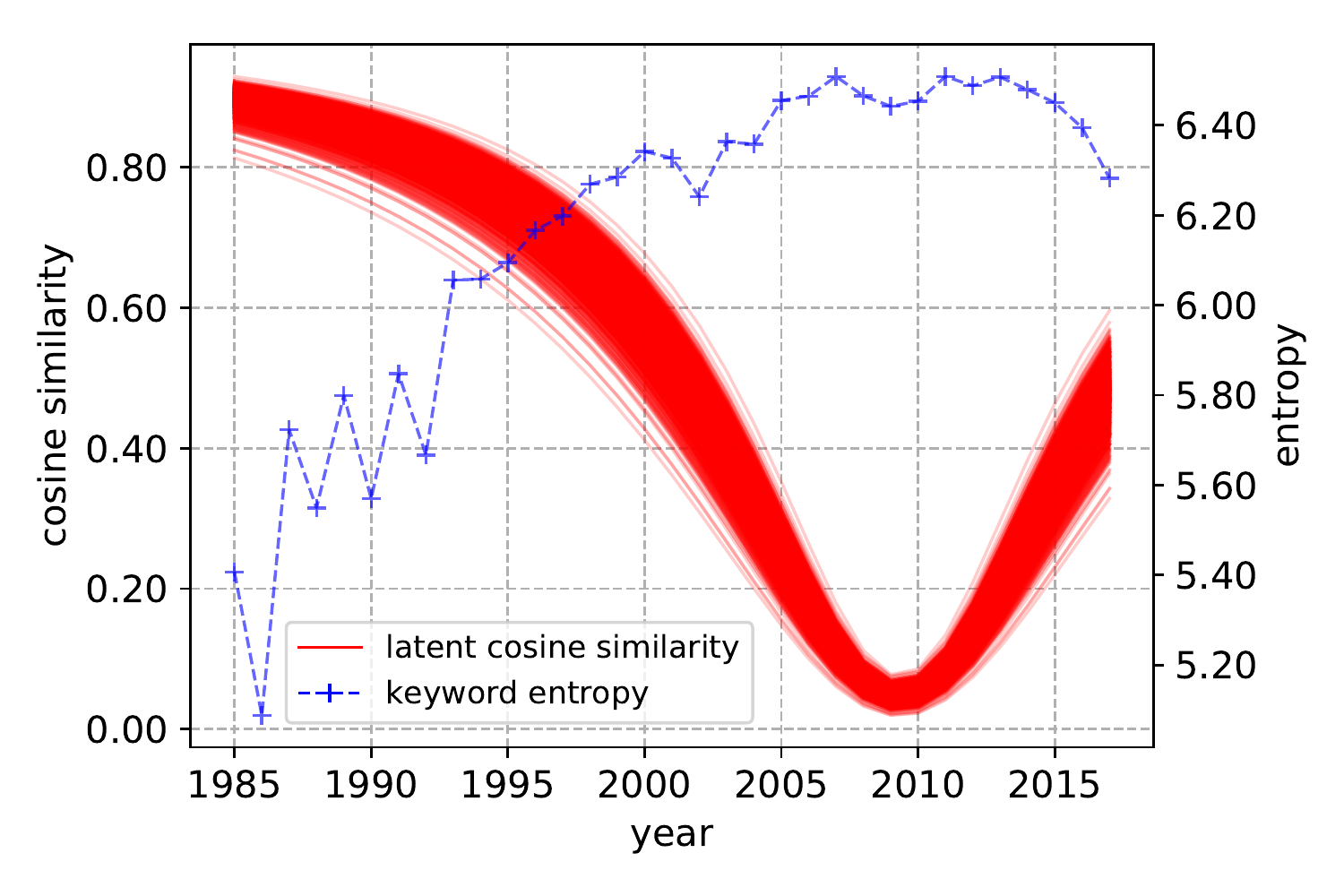}
\caption{Evolution of latent vectors. Red lines correspond to the averaged cosine similarity between authors in the latent space in the S2 corpus. The blue dotted line is the entropy of keywords at each timestep.}
\label{fig:s2-sim-ent}
\end{figure}

We further analyze the learned trajectories on S2 by examining cosine similarities between authors in the latent space. We show on \figref{fig:s2-sim-ent} the average cosine similarity between authors through time. First, we see that all authors follow the same trend. It was expected since we saw so in \figref{fig:latent-s2-sc} that all authors seem to follow the same dynamics. On the first timesteps, all representations are very similar, with a cosine similarity around $0.9$. Since there are only a few documents published at these timesteps, and because of the weight decay on $h_{a}$, all representations tend to regroup in the same place, preventing overfitting. The average similarity then drops to $0$, as the model learns to drive away each representation to better fit them to each author. And then, after 2009, the average similarities go up to and reach $0.5$ on the last timestep. This sudden augmentation in global similarity cannot be explained by the quantity of data, as the last 6 timesteps contain $50\%$ of the documents in the corpus. Another hypothesis is that global diversity among authors diminishes. To illustrate this, we plot the entropy of articles' keywords through time, that we interpret as the diversity of subjects studied in the community. The entropy is plotted in blue on \figref{fig:s2-sim-ent}, and we can see that, symmetrically to the cosine similarities, the entropy of keywords increases from 1985 to 2010 approximately, and then begins to drop. This drop of entropy indicates that the diversity of topics also drops, and is translated by our model in an augmentation of the average similarity between authors.

\subsection{Latent Space Analysis: NYT}
\label{sec:xp-lat-nyt}

\begin{figure}[t]
\centering
\includegraphics[width=0.8\linewidth]{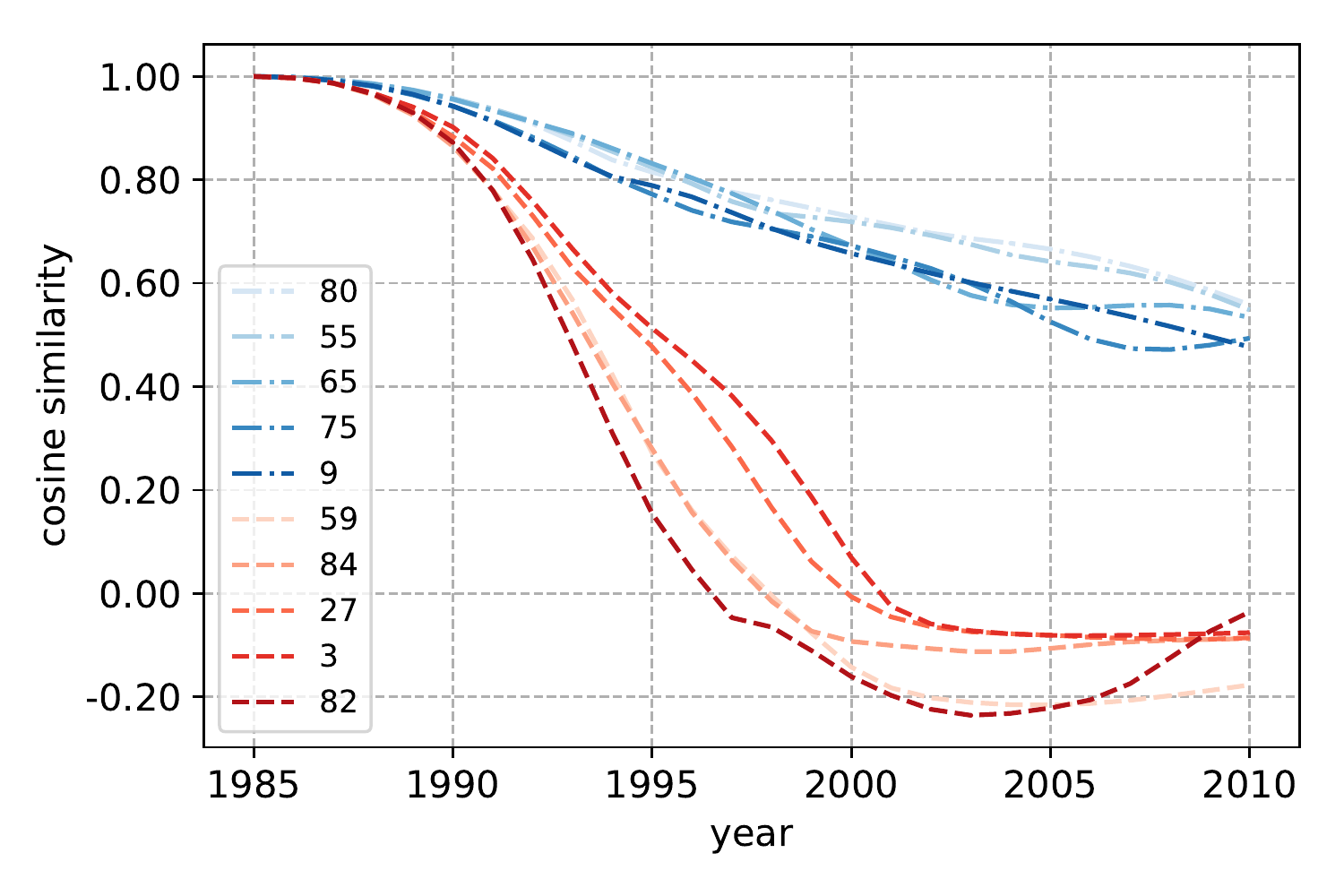}
\caption{Cosine similarity between $h_{a, 1}$ and $h_{a, t}$ for all $t$. Blue lines (top) are authors $a$ with the highest cosine similarity between $h_{a, 1}$ and $h_{a, T}$, and red lines (bottom) are those with the smallest. The legend indicates authors id.}
\label{fig:nyt-self-sim}
\end{figure}

\begin{table}
    \caption{Evolution of labels for the 5 authors that change the least (top) and that change the most (bottom) in the latent space on the NYT corpus. The numbers in italic font at the top of each column are author ids, they match with the ids in the legend of \figref{fig:nyt-self-sim}.}
    \centering
    \label{tab:nyt-cl-evo}
    \begin{tabular}{|r||l|l|l|l|l|}
        \hline
        \multicolumn{1}{|c||}{} & \multicolumn{5}{c|}{\textbf{Least changes}} \\
        \hline
        \multicolumn{1}{|c||}{\textit{year}} & \multicolumn{1}{c|}{\textit{80}} & \multicolumn{1}{c|}{\textit{55}} & \multicolumn{1}{c|}{\textit{65}} & \multicolumn{1}{c|}{\textit{75}} & \multicolumn{1}{c|}{\textit{9}} \\
        \hline
        1991 & N.Y.C. & home & sports & front page & business \\
        1993 & N.Y.C. & home & sports & front page & N.Y.C. \\
        1995 & N.Y.C. & home & sports & U.S. & U.S. \\
        1997 & N.Y.C. & home & sports & U.S. & U.S. \\
        1999 & N.Y.C. & dining & sports & U.S. & U.S. \\
        2001 & N.Y.C. & dining & sports & front page & U.S. \\
        2003 & N.Y.C. & dining & sports & U.S. & U.S. \\
        2005 & N.Y.C. & dining & - & U.S. & U.S. \\
        2007 & N.Y.C. & arts & - & washington & business day \\
        2009 & N.Y.C. & food & - & U.S. & washington \\
        2011 & N.Y.C. & dining & - & - & science \\
        2013 & N.Y.C. & dining & - & opinion & U.S. \\
        2015 & N.Y.C. & food & - & opinion & business day \\
        \hline
        \hline
        \multicolumn{1}{|c||}{} & \multicolumn{5}{c|}{\textbf{Most changes}} \\
        \hline
        \multicolumn{1}{|c||}{\textit{year}} & \multicolumn{1}{|c|}{\textit{59}} & \multicolumn{1}{c|}{\textit{84}} & \multicolumn{1}{c|}{\textit{27}} & \multicolumn{1}{c|}{\textit{3}} & \multicolumn{1}{c|}{\textit{82}} \\
        \hline
        1991 & N.Y.C. & N.Y.C. & world & arts & - \\
        1993 & N.Y.C. & N.Y.C. & U.S. & movies & - \\
        1995 & arts & front page & U.S. & arts & - \\
        1997 & arts & U.S. & U.S. & arts & - \\
        1999 & world & U.S. & world & dining & N.Y.C. \\
        2001 & world & U.S. & world & dining & N.Y.C. \\
        2003 & world & world & world & dining & arts \\
        2005 & world & U.S. & world & arts & movies \\
        2007 & world & U.S. & world & books & theater \\
        2009 & world & - & world & arts & arts \\
        2011 & world & - & - & arts & arts \\
        2013 & world & - & world & travel & arts \\
        2015 & sports & - & U.S. & books & arts \\
        \hline
    \end{tabular}
\end{table}

On the NYT corpus, contrary to S2, we notice that our model learns different dynamics depending on authors (\figref{fig:latent-nyt-sc}). We explore two modes of variations by analyzing authors that change the most, and authors that change the least between 1985 and 2015. We measure change with the cosine similarity between first and last latent representations $h_{a, 1}$ and $h_{a, T}$. We restrict our study on the 100 (out of 500) authors that published the most in order to reduce noise. On \figref{fig:nyt-self-sim} we plot the cosine similarity between $h_{a, 1}$ and $h_{a, t})$ at all timesteps for the 5 authors that change the most and the 5 that change the least. We can see that the dynamics are different for the two groups, one decreasing slowly to $0.6$ of cosine similarity, and the other dropping faster to $0$.

We now want to assess that those differences in dynamics in the latent space are linked to the author's publications evolution. To do so, we use the venue of the articles. Venues can be viewed as broad subjects attached to each article. We only select those appearing at least 50 times in the dataset, which gives us 31 topics. We associate to each ($a$, $t$) couple the most represented topic in the articles published by $a$ at $t$.

\begin{table*}[t]
    \centering
    \caption{Samples from our model conditioned on different authors through time. Text sequences are generated by feeding the first three word displayed in bold at the top of each. The samples were obtained by beam search with a beam size of 5.}
    \label{tab:s2-generation}
    \begin{tabular}{|l|l|l|l|}
        \hline
        \multicolumn{1}{|}{} & \multicolumn{1}{|c|}{1} & \multicolumn{1}{c|}{2} & \multicolumn{1}{c|}{3} \\
        \hline
        \hline
        \multicolumn{1}{|c}{A} & \multicolumn{3}{|c|}{\textbf{semi - supervised}...}\\
        \hline
        1985 & ...learning & ...learning in the presence of noise & ...learning of object categories \\
        1990 & ...learning with the em algorithm & ...learning of linear models & ...learning of object categories \\
        1995 & ...learning with a probabilistic model & ...learning of probabilistic models & ...image segmentation \\
        2000 & ...learning with kernels & ...learning with gaussian processes & ...segmentation of 3d objects \\
        2005 & ...learning with pairwise constraints & ...learning for text classification & ...segmentation of 3d human motion \\
        2010 & ...learning with pairwise constraints & ...learning for text classification & ...multi - view face recognition \\
        2015 & ...learning with deep neural networks & ...multi - task learning & ...convolutional neural networks \\
        2016 & ...learning with deep neural networks & ...learning with deep neural networks & ...convolutional neural networks \\
        2017 & ...learning with deep neural networks & ...deep learning & ...convolutional neural networks \\

        \hline
        \hline
        \multicolumn{1}{|c}{B} & \multicolumn{3}{|c|}{\textbf{a study of}...}\\
        \hline
        1985 & ...image segmentation & ...knowledge compilation & ...word sense disambiguation \\
        1990 & ...the fundamental matrix & ...knowledge compilation & ...word sense disambiguation \\
        1995 & ...multi - view stereo & ...bayesian networks & ...statistical machine translation \\
        2000 & ...image segmentation algorithms & ...probabilistic logic programming & ...statistical machine translation \\
        2005 & ...multi - view face recognition & ...probabilistic models for relational learning & ...statistical machine translation \\
        2010 & ...energy minimization algorithms & ...probabilistic models for relational learning & ...statistical machine translation \\
        2015 & \shortstack{...modern inference techniques for \\ structured prediction}  & ...variational bayesian inference & ...statistical machine translation systems \\
        2016 & \shortstack{...modern inference techniques for \\ structured prediction} & ...variational bayesian inference & ...neural machine translation systems \\
        2017 & ...deep convolutional neural networks & ...variational bayesian inference & ...neural machine translation systems \\

        \hline
        \hline
        \multicolumn{1}{|c}{C} & \multicolumn{3}{|c|}{\textbf{real - time}...}\\
        \hline
        1985 & ...visual tracking & ...visual tracking & ...visualization of the web \\
        1990 & ...visual tracking & ...visual tracking & ...multi - view stereo \\
        1995 & ...visual tracking & ...visual tracking & ...time - series classification \\
        2000 & ...multi - view clustering & ...visual tracking & ...time series classification \\
        2005 & ...collaborative filtering & ...facial expression recognition & ...time series classification \\
        2010 & ...bidding in display advertising & ...visual tracking using deep learning & ...time series classification \\
        2015 & ...bidding in display advertising & ...visual tracking with deep neural networks & ...time series forecasting \\
        2016 & ...bidding in display advertising & ...facial expression recognition & ...time series forecasting \\
        2017 & ...bidding in display advertising & ...visual tracking with deep neural networks & ...time series forecasting \\

        \hline
        \hline
        \multicolumn{1}{|c}{D} & \multicolumn{3}{|c|}{\textbf{a framework for}...}\\
        \hline
        1985 & ...learning to rank & ...qualitative simulation & ...learning to rank \\
        1990 & ...learning to rank & ...multi - agent reinforcement learning & ...learning to rank \\
        1995 & ...learning to rank & ...multi - agent reinforcement learning & ...learning to rank \\
        2000 & ...parsing natural language & ...multi - agent reinforcement learning & ...learning to rank \\
        2005 & ...parsing natural language & ...multi - agent reinforcement learning & ...learning to rank \\
        2010 & ...multi - task learning & ...multi - target tracking & ...learning to rank \\
        2015 & ...multi - task learning & ...multi - target tracking & ...learning to rank \\
        2016 & ...multi - task learning & ...multi - target tracking & ...learning to rank \\
        2017 & ...recurrent neural networks & ...deep reinforcement learning & ...learning to rank \\

        \hline
    \end{tabular}
\end{table*}

We now have a label for each author/timestep couple, if at least one article is published. We show in table \ref{tab:nyt-cl-evo} the evolution of topics for the authors plotted in \figref{fig:nyt-self-sim}. A dash (-) correspond to a year without publication. We slightly change the topics' names to fit the table into the paper. On top, we have authors that change the least in the latent space, and we see that their topics are the same, or very close semantically. For instance \texttt{dinning}, \texttt{food}, and \texttt{home} all correspond to lifestyle venues. Only author 9 has some varying topics in the last years. On the lower half of the table, we have authors that change the most. We see that each of them has at least two significantly different topics: \texttt{arts} and \texttt{world} for 59, \texttt{N.Y.C.} and \texttt{world} for 84, \texttt{U.S.} and \texttt{world} for 27, \texttt{movies} and \texttt{dinning} for 3, and \texttt{N.Y.C} and \texttt{arts} for 82. This indicates that latent dynamics learned by our model are indeed observable in the document space.

\subsection{Data Samples}
\label{sec:xp-gen}

Language models are generative models, and it is thus possible to sample text from them. Here, we present samples generated by our model trained on Semantic Scholar for the modeling task. Each sample is generated by beam search with a beam of size 5, and is seeded with different word triplets that often appear in the corpus.

We conditioned the LSTM decoder of our model to authors randomly sampled, at several timesteps. The samples are presented in table \ref{tab:s2-generation}. Each box from A to D corresponds to a word triplet seed and each column from 1 to 3 to an author. Note that authors are different between blocks, and the author of A1 is not the author of B1.

We first notice that samples are smooth in time in the text space. We also see different speeds of variation between the different authors. For instance, the samples D3 (bottom right) are always exactly the same at each timestep, while the B1 samples vary rapidly. Generally, we can see that our model tends to generate titles related to deep neural networks at the last timesteps of every author (\texttt{recurrent neural networks}, \texttt{deep reinforcement learning}, \texttt{deep convolutional neural networks}, etc...). It is consistent with the increase in average author similarity found in \secref{sec:xp-lat-s2}. We also see that samples for a particular author across time tend to refer to the same sub-field (e.g. computer vision or natural language processing), which is also consistent to dynamics observed in \secref{sec:xp-lat-s2}.

\section{Conclusion}
Modes of expression in author communities evolve over time because of internal or external factors. It thus appears crucial to be able to capture these dynamics, as much for analysis as for language modeling tasks. In this paper, we proposed a model that seeks to fill this identified need, by leveraging the recent advances in representation learning and neural networks. The proposed model aims at capturing the evolution dynamics of language in author communities, by exploiting dependencies between successive steps. We modeled each author by a representation vector that evolves dynamically in time with a residual function conditioned on a static author representation. Experimental results show that the proposed model improves  modeling, imputation, and prediction of language distributions in author communities.

In future works, we are interested in explicitly discovering relationships between authors. The proposed method has the potential to capture relations in the latent space, but only implicitly. In order to fully address language diffusion problems, we are currently working on a relational extension. This would allow us to explicitly capture relations between authors, and study their evolution through time. Recently, a new kind of LM architecture based on transformer networks \cite{devlin18, radford2019language} achieved state of the art results in various NLP tasks. Integrating it and analyzing its effects in our framework is an interesting and promising research direction.

\bibliographystyle{IEEEtran}
\bibliography{biblio}

\end{document}